\documentclass[11pt,a4paper]{article}
\usepackage[hyperref]{emnlp-ijcnlp-2019}
\usepackage{times}
\usepackage{url}
\usepackage{latexsym}
\usepackage[utf8]{inputenc}
\usepackage{amsmath}
\usepackage{amsfonts}
\usepackage{booktabs}
\usepackage{algorithm}
\usepackage[noend]{algpseudocode}
\usepackage{comment}
\usepackage{bbm}
\usepackage{natbib}
\usepackage{stmaryrd}
\usepackage{empheq}
\usepackage{mathtools}
\usepackage{graphicx,multirow}
\usepackage[normalem]{ulem}
\usepackage{subcaption}
\usepackage{siunitx}
\usepackage{enumitem}
\usepackage{pifont}

\newcommand{\joeynmt}{Joey NMT }
\newcommand{\cmark}{\ding{51}}
\newcommand{\xmark}{\ding{55}}

\aclfinalcopy 
\def\aclpaperid{40} 

\title{Joey NMT: A Minimalist NMT Toolkit for Novices}

\author{Julia Kreutzer \\
  Computational Linguistics \\
  Heidelberg University \\
  {\tt \small kreutzer@cl.uni-heidelberg.de} \\\And
  Jasmijn Bastings \\
  ILLC \\
University of Amsterdam \\
  {\tt \small bastings@uva.nl} \\\And
  Stefan Riezler\\
  Computational Linguistics \& IWR \\
Heidelberg University \\
 {\tt \small riezler@cl.uni-heidelberg.de}
  }

\date{}

\begin{document}
\maketitle

\begin{abstract}
We present Joey NMT, a minimalist neural machine translation toolkit based on PyTorch that is specifically designed for novices. \joeynmt provides many popular NMT features in a small and simple code base, so that novices can easily and quickly learn to use it and adapt it to their needs. Despite its focus on simplicity, \joeynmt supports classic architectures (RNNs, transformers), fast beam search, weight tying, and more, and achieves performance comparable to more complex toolkits on standard benchmarks.
We evaluate the accessibility of our toolkit in a user study where novices with general knowledge about Pytorch and NMT and experts work through a self-contained \joeynmt tutorial, showing that novices perform almost as well as experts in a subsequent code quiz. 
\joeynmt is available at \url{https://github.com/joeynmt/joeynmt}.
\end{abstract}

\section{Introduction}

Since the first successes of neural machine translation (NMT), various research groups and industry labs have developed open source toolkits specialized for NMT, based on new open source deep learning platforms.  While toolkits like OpenNMT \cite{OpenNMT}, XNMT \cite{XNMT} and Neural Monkey \cite{NeuralMonkey:2017} aim at readability and extensibility of their codebase, their target group are researchers with a solid background in machine translation and deep learning, and with experience in navigating, understanding and handling large code bases. However, none of the existing NMT tools has been designed primarily for readability or accessibility for novices, nor has anyone studied quality and accessibility of such code empirically.  
On the other hand, it is an important challenge for novices to understand how NMT is implemented, what features each toolkit implements exactly, and which toolkit to choose in order to code their own project as fast and simple as possible.

We present an NMT toolkit especially designed for novices, providing clean, well documented, and minimalistic code, that is yet of comparable quality to more complex codebases on standard benchmarks. Our approach is to identify the core features of NMT that have not changed over the last years, and to invest in  documentation, simplicity and quality of the code. These core features include standard network architectures (RNN, transformer, different attention mechanisms, input feeding, configurable encoder/decoder bridge), standard learning techniques (dropout, learning rate scheduling, weight tying, early stopping criteria), and visualization/monitoring tools.

We evaluate our codebase in several ways: Firstly, we show that Joey NMT's comment-to-code ratio is almost twice as high as other toolkits which are roughly 9-10 times larger. Secondly, we present an evaluation on standard benchmarks (WMT17, IWSLT) where we show that the core architectures implemented in Joey NMT achieve comparable performance to more complex state-of-the-art toolkits. Lastly, we conduct a user study where we test the code understanding of novices, i.e. students with basic knowledge about NMT and PyTorch, against expert coders. While novices, after having worked through a self-contained \joeynmt tutorial, needed more time to answer each question in an in-depth code quiz, they achieved only marginally lower scores than the experts. To our knowledge, this is the first user study on the accessibility of NMT toolkits.

\section{\joeynmt}\label{sec:\joeynmt}

\subsection{NMT Architectures}\label{sec:models}

This section formalizes the \joeynmt implementation of autoregressive recurrent and fully-attentional models.

In the following, a source sentence of length $l_x$ is represented by a sequence of one-hot encoded vectors $\mathbf{x}_1, \mathbf{x}_2, \dots, \mathbf{x}_{l_x}$ for each word.
Analogously, a target sequence of length $l_y$ is represented by a sequence of one-hot encoded vectors $\mathbf{y}_1, \mathbf{y}_2, \dots, \mathbf{y}_{l_y}$.

\subsubsection{RNN}
\joeynmt implements the RNN encoder-decoder variant from \citet{LuongETAL:15}.

\paragraph{Encoder.} The encoder RNN transforms the input sequence $\mathbf{x}_1, \dots, \mathbf{x}_{l_x}$ into a sequence of vectors $\mathbf{h}_1, \dots, \mathbf{h}_{l_x}$ with the help of the embeddings matrix $E_{src}$ and a recurrent computation of states
\begin{align*}
    \mathbf{h}_i &= \text{RNN}(E_{src}\, \mathbf{x}_i, \mathbf{h}_{i-1}); &\mathbf{h}_0 = \mathbf{0}.
\end{align*}
The RNN consists of either GRU or a LSTM units. For a bidirectional RNN, hidden states from both directions are are concatenated to form $\mathbf{h}_i$. The initial encoder hidden state $\mathbf{h}_0$ is a vector of zeros. 
Multiple layers can be stacked by using each resulting output sequence $\mathbf{h}_1, \dots, \mathbf{h}_{l_x}$ as the input to the next RNN layer.

\paragraph{Decoder.} 
The decoder uses input feeding \cite{LuongETAL:15} where an attentional vector  $\mathbf{\tilde{s}}$ is concatenated with the representation of the previous word as input to the RNN. Decoder states are computed as follows:
\begin{align*}
    \mathbf{s}_t &= \text{RNN}([E_{trg}\, \mathbf{y}_{t-1}; \mathbf{\tilde{s}}_{t-1}], \mathbf{s}_{t-1})\\
   \mathbf{s}_0 &= \begin{cases} \tanh(W_{bridge}\, \mathbf{h}_{l_x} + \mathbf{b}_{bridge}) & \text{if bridge} \\
                                \mathbf{h}_{l_x} & \text{if last} \\
                                \mathbf{0} & \text{otherwise}
                                \end{cases} \\
    \mathbf{\tilde{s}}_{t} &= \tanh(W_{att} [\mathbf{s}_{t}; \mathbf{c}_{t}] + \mathbf{b}_{att})
\end{align*}
The initial decoder state is configurable to be either a non-linear transformation of the last encoder state (``bridge''), or identical to the last encoder state (``last''), or a vector of zeros.

\paragraph{Attention.} 
The context vector $\mathbf{c}_t$ is computed with an attention mechanism scoring the previous decoder state $\mathbf{s}_{t-1}$ and each encoder state $\mathbf{h}_i$:
\begin{align*}
    \mathbf{c}_t &= \sum_{i} a_{ti} \cdot \mathbf{h}_i\\
    a_{ti} &= \frac{\exp(\text{score}(\mathbf{s}_{t-1}, \mathbf{h}_i))}{\sum_{k} \exp(\text{score}(\mathbf{s}_{t-1}, \mathbf{h}_k))}
\end{align*}
where the scoring function is a multi-layer perceptron \cite{BahdanauETAL:15} or a bilinear transformation \cite{LuongETAL:15}.

\paragraph{Output.}
The output layer produces a vector 
$\mathbf{o}_t = W_{out}\, \mathbf{\tilde{s}}_t$, 
which contains a score for each token in the target vocabulary.
Through a softmax transformation, these scores can be interpreted as a probability distribution over the target vocabulary $\mathcal{V}$ that defines an index over target tokens $v_j$.
\begin{align*}
    p(y_t = v_j \mid x, y_{<t}) = \frac{\exp(\mathbf{o}_t[j])}{\sum_{k=1}^{|\mathcal{V}|}\exp(\mathbf{o}_t[k])}
\end{align*}

\subsubsection{Transformer}\label{sec:transformer}
\joeynmt implements the Transformer from \citet{vaswani2017attention}, with code based on \emph{The Annotated Transformer} blog \citep{rush2018annotated}.

\paragraph{Encoder.} Given an input sequence $\mathbf{x}_1, \dots, \mathbf{x}_{l_x}$, we look up the word embedding for each input word using $E_{src} \mathbf{x}_i$, add a position encoding to it, and stack the resulting sequence of word embeddings to form matrix $X \in \mathbb{R}^{l_x \times d}$, where $l_x$ is the sentence length and $d$ the dimensionality of the embeddings. 

We define the following learnable parameters:\footnote{Exposition adapted from Michael Collins \url{https://youtu.be/jfwqRMdTmLo}}
$$
 A \in \mathbb{R}^{d \times d_a} \quad	B \in \mathbb{R}^{d \times d_a}	\quad C \in \mathbb{R}^{d \times d_o}
$$
where $d_a$ is the dimensionality of the attention (inner product) space and $d_o$ the output dimensionality.
Transforming the input matrix with these matrices into new word representations $H$
$$
H = \underbrace{\text{softmax}\big( X\!A \, B^\top \!\! X^\top \big)}_{\text{self-attention}} \, X\!C
$$
which have been updated by attending to all other source words. \joeynmt implements multi-headed attention, where this transformation is computed $k$ times, one time for each head with different parameters $A, B, C$.

After computing all $k$ $H$s in parallel, we concatenate them and apply layer normalization and a final feed-forward layer:
\begin{align*}
H  &= [ H^{(1)}; \dots ; H^{(k)} ] \\
H'  &= \text{layer-norm}(H) + X\\
H^{\text{(enc)}} &= \text{feed-forward}(H') + H'
\end{align*}
We set $d_o = d / k$, so that $H \in \mathbb{R}^{l_x \times d}$. Multiple of these layers can be stacked by setting $X=H^{\text{(enc)}}$ and repeating the computation.
\paragraph{Decoder.} The Transformer decoder operates in a similar way as the encoder, but takes the stacked target embeddings $Y\!\!\in\!\!\mathbb{R}^{l_y \times d}$ as input:
$$
H = \underbrace{\text{softmax}\big( Y\!A \, B^{\top}\!Y^{\top}\big)}_{\text{masked self-attention}} Y\!C 
$$
For each target position attention to future input words is inhibited
by setting those attention scores to $-inf$ before the $\text{softmax}$. 
After obtaining $H' = H + Y$, and before the feed-forward layer, we compute multi-headed attention again, 
but now between intermediate decoder representations $H'$ and final encoder representations $H^{\text{(enc)}}$:
\begin{align*}
Z &= \underbrace{\text{softmax}\big( H'A \, B^\top {H^{\text{(enc)}}}^\top \big)}_{\text{src-trg attention}} \, H^{\text{(enc)}}C \\
H^{\text{(dec)}} & = \text{feed-forward}(\text{layer-norm}(H' + Z))
\end{align*}
We predict target words with $H^{\text{(dec)}}W_{out}$.

\subsection{Features}\label{sec:features}

In the spirit of minimalism, we follow the 80/20 principle \cite{pareto1896cours} and aim to achieve 80\% of the translation quality
with 20\% of a common toolkit's code size. For this purpose we identified the most common features (the bare necessities) in recent works and implementations.\footnote{We refer the reader to the additional technical description in \url{https://arxiv.org/abs/1907.12484}: Table 6
in Appendix A.1
compares Joey NMT's features with several popular NMT toolkits and shows that \joeynmt covers all features that those toolkits have in common.}
It includes standard architectures (see §\ref{sec:models}), label smoothing, dropout in multiple places, various attention mechanisms, input feeding, configurable encoder/decoder bridge, learning rate scheduling, weight tying, early stopping criteria, beam search decoding, an interactive translation mode, visualization/monitoring of learning progress and attention, checkpoint averaging, and more. 

\subsection{Documentation}\label{sec:documentation}
The code itself is documented with doc-strings and in-line comments (especially for tensor shapes), and modules are tested with unit tests. The documentation website\footnote{\url{https://joeynmt.readthedocs.io}} contains installation instructions, a walk-through tutorial for training, tuning and testing an NMT model on a toy task\footnote{Demo video: \url{https://youtu.be/PzWRWSIwSYc}}, an overview of code modules, and a detailed API documentation. In addition, we provide thorough answers to frequently asked questions regarding usage, configuration, debugging, implementation details and code extensions, and recommend resources, such as data collections, PyTorch tutorials and NMT background material. 

\subsection{Code Complexity}\label{sec:codecomplexity}

In order to facilitate fast code comprehension and navigation \cite{wiedenbeck1999comparison}, \joeynmt objects have at most one level of inheritance. 
Table~\ref{tab:cloc} compares \joeynmt with OpenNMT-py and XNMT (selected for their extensibility and thoroughness of documentation) in terms of code statistics, i.e. lines of Python code, lines of comments and number of files.\footnote{Using \url{https://github.com/AlDanial/cloc}} 
OpenNMT-py and XNMT have roughly 9-10x more lines of code, spread across 4-5x more files than \joeynmt. These toolkits cover more than the essential features for NMT (see §\ref{sec:features}), in particular for other generation or classification tasks like image captioning and language modeling. However, Joey NMT's comment-to-code ratio is almost twice as high, which we hope will give code readers better guidance in understanding and extending the code. 

\begin{table}[t]
    \centering
    \resizebox{\columnwidth}{!}{
    \begin{tabular}{l|ccc}
        \toprule
        \textbf{Counts} & \textbf{OpenNMT-py} &\textbf{XNMT} & \textbf{\joeynmt} \\
        \midrule
        Files &  94 & 82 & 20 \\
        Code & 10,287 & 11,628 & 2,250 \\ 
        Comments & 3,372 & 4,039 & 1,393\\ 
        \midrule 
        Comment/Code Ratio & 0.33 & 0.35 & \textbf{0.62}\\
        \bottomrule
    \end{tabular}
    }
    \caption{Python code statistics for OpenNMT-py (commit hash \texttt{624a0b3a}), XNMT (\texttt{a87e7b94}) and \joeynmt (\texttt{e55b615}).}
    \label{tab:cloc}

\end{table}

\subsection{Benchmarks}\label{sec:benchmarks}

Our goal is to achieve a performance that is comparable to other NMT toolkits, so that novices can start off with reliable benchmarks that are trusted by the community. This will allow them to build on \joeynmt for their research, should they want to do so.
We expect novices to have limited resources available for training, i.e., not more than one GPU for a week, and therefore we focus on benchmarks that are within this scope. Pre-trained models, data preparation scripts and configuration files for the following benchmarks will be made available on \url{https://github.com/joeynmt/joeynmt}. 
\begin{table*}[!th]
    \centering
    \begin{tabular}{lccccccc}
        \toprule
        \multirow{2}{*}{\textbf{System}} & \multicolumn{2}{c}{\textbf{Groundhog RNN}} & \multicolumn{3}{c}{\textbf{Best RNN}}& \multicolumn{2}{c}{\textbf{Transformer}} \\
        \cmidrule(lr){2-3} \cmidrule(lr){4-6} \cmidrule(lr){7-8}
        & \textbf{en-de} & \textbf{lv-en} & \textbf{layers} &\textbf{en-de} & \textbf{lv-en} & \textbf{en-de} & \textbf{lv-en} \\
        \midrule
         NeuralMonkey       & 13.7    & 10.5 & 1/1       & 13.7      & 10.5 & -- & -- \\
         OpenNMT-Py         & 18.7    & 10.0 & 4/4       & 22.0      & 13.6 & -- & --\\
         Nematus            & 23.9    & 14.3  & 8/8      & 23.8      & 14.7 & -- & --\\
         Sockeye            & 23.2    & 14.4 &  4/4       & 25.6      & 15.9 & 27.5 & 18.1 \\
         Marian             & 23.5    & 14.4  &  4/4    & 25.9      & 16.2 & 27.4 & 17.6 \\
         Tensor2Tensor      &  --       &  --     &     --     &    --     & -- & 26.3 & 17.7 \\
         \midrule
         \textbf{\joeynmt} & 23.5 & 14.6  & 4/4 & 26.0       & 15.8  & 27.4 & 18.0 \\
        \bottomrule
    \end{tabular}
    \caption{Results on WMT17 \texttt{newstest2017}. Comparative scores are from \citet{hieber2018sockeye}.}
    \label{tab:wmt17all}
\end{table*}

\paragraph{WMT17.}
We use the settings of \citet{hieber2018sockeye}, using the exact same data, pre-processing, and evaluation using WMT17-compatible SacreBLEU scores \citep{sacrebleu}.\footnote{\tiny\path{BLEU+case.mixed+lang.[en-lv|en-de]+numrefs.1+smooth.exp+test.wmt17+tok.13a+version.1.3.6}}
We consider the setting where toolkits are used out-of-the-box to train a Groundhog-like model (1-layer LSTMs, MLP attention), the `best found' setting where \citeauthor{hieber2018sockeye} train each model using the best settings that they could find, and the Transformer base setting.\footnote{Note that the scores reported for other models reflect their state when evaluated in \citet{hieber2018sockeye}.}
 Table~\ref{tab:wmt17all} shows that \joeynmt performs very well compared against other shallow, deep and Transformer models, despite its simple code base.\footnote{Blog posts like \citet{rush2018annotated} and \citet{bastings2018annotated} also offer simple code, but they do not perform as well.}

\paragraph{IWSLT14.} This is a popular benchmark because of its relatively small size and therefore fast training time.
We use the data, pre-processing, and word-based vocabulary of \citet{wiseman-rush-2016-sequence} and evaluate with SacreBLEU.\footnote{\tiny\path{BLEU+case.lc+numrefs.1+smooth.exp+tok.none+version.1.3.6}}
Table \ref{tab:iwslt-de} shows that \joeynmt performs well here, with both its recurrent and its Transformer model.
We also included BPE results for future reference.

\begin{table}[ht]
    \centering
    \begin{tabular}{lc}
    \toprule
        \textbf{System} & \textbf{de-en} \\
    \midrule
        \citet{wiseman-rush-2016-sequence}  &  22.5 \\ 
        \citet{BahdanauETAL:16} & 27.6 \\ 
        \textbf{\joeynmt} (RNN, word) & 27.1	\\
        \textbf{\joeynmt} (RNN, BPE32k) & 27.3	\\
        \textbf{\joeynmt} (Transformer, BPE32k) & 31.0 \\ 
    \bottomrule
    \end{tabular}
    \caption{IWSLT14 test results.}
    \label{tab:iwslt-de}
\end{table}

\section{User Study}
The target group for \joeynmt are novices who will use NMT in a seminar project, a thesis, or an internship. Common tasks are to re-implement a paper, extend standard models by a small novel element, or to apply them to a new task.
In order to evaluate how well novices understand Joey NMT, we conducted a user study comparing the code comprehension of novices and experts.

\subsection{Study Design}
\paragraph{Participants.}
The novice group is formed of eight undergraduate students with a Computational Linguistics major that have all passed introductory courses to Python and Machine Learning, three of them also a course about Neural Networks. 
None of them had practical experience with training or implementing NMT models nor PyTorch, but two reported theoretic understanding of NMT. 
They attended a 20h crash course introducing NMT and Pytorch basics.\footnote{See §\ref{sec:course} in the supplemental material of \url{https://arxiv.org/abs/1907.12484} for details.}
Note that we did not teach \joeynmt explicitly in class, but the students independently completed the \joeynmt tutorial.

As a control group (the ``experts''), six graduate students with NMT as topic of their thesis or research project participated in the study. In contrast to the novices, this group of participants has a solid background in Deep Learning and NMT, had practical experience with NMT.  All of them had previously worked with NMT in PyTorch.

\paragraph{Conditions.} 
The participation in the study was voluntary and not graded. Participants were not allowed to work in groups and had a maximum time of 3h to complete the quiz. They had previously locally installed Joey NMT\footnote{\joeynmt commit hash \texttt{0708d596}, prior to the Transformer implementation.} and could browse the code with the tools of their choice (IDE or text editor).  They were instructed to explore the \joeynmt code with the help of the quiz, informed about the purpose of the study, and agreed to the use of their data in this study. Both groups of participants had to learn about \joeynmt in a self-guided manner, using the same tutorial, code, and documentation.  
The quiz was executed on the university's internal e-learning platform. Participants could jump between questions, review their answers before finally submitting all answers and could take breaks (without stopping the timer).
Answers to the questions were published after all students had completed the test.

\paragraph{Question design.}
The questions are not designed to test the participant's prior knowledge on the topic, but to guide their exploration of the code. The questions are either free text, multiple choice or binary choice.
There are three blocks of questions:\footnote{\url{https://arxiv.org/abs/1907.12484} contains the full list of questions, complete statistics and details of the LME analysis.}
\begin{enumerate}
    \item \textbf{Usage of \joeynmt}: nine questions on how to interpret logs, check whether models were saved, interpret attention matrices, pre-/post-process, and to validate whether the model is doing what it is built for.
    \item \textbf{Configuring \joeynmt}: four questions that make the users configure \joeynmt in such a way that it works for custom situations, e.g. with custom data, with a constant learning rate, or creating model of desired size.
    \item \textbf{\joeynmt Code}: eighteen questions targeting the detailed understanding of the \joeynmt code: the ability to navigate between python modules, identify dependencies, and interpret what individual code lines are doing, hypothesize how specific lines in the code would have to get changed to change the behavior (e.g. working with a different optimizer). The questions in this block were designed in a way that in order to find the correct answers, every python module contained in \joeynmt had to be visited at least once.
\end{enumerate}

Every question is awarded one point if answered correctly. Some questions require manual grading, most of them have one correct answer. We record overall completion time and time per question.\footnote{Time measurement is noisy, since full minutes are measured and students might take breaks at various points in time.}

\subsection{Analysis}
\paragraph{Total duration and score.} Experts took on average 77 min to complete the quiz, novices 118 min, which is significantly slower (one-tailed t-test, $p<0.05$). 
Experts achieved on average 82\% of the total points, novices 66\%.
According to the t-test the difference in total scores between groups is significant at $p<0.05$. 
An ANOVA reveals that there is a significant difference in total duration and scores within the novices group, but not within the experts group.

\paragraph{Per question analysis.} No question was incorrectly answered by everyone. Three questions (\#6, \#11, \#18) were correctly answered by everyone--they were appeared to be easiest to answer and did not require deep understanding of the code. 
In addition, seven questions (\#1, \#13, \#15, \#21, \#22, \#28, \#29) were correctly answered by all experts, but not all novices--here their NMT experience was useful for working with hyperparameters and peculiarities like special tokens.
However, for only one question, regarding the differences in data processing between training and validation (\#16), the difference between average expert and novice score was significant (at $p < 0.05$). Six questions (\#9, \#18, \#21, \#25, \#31) show a significantly longer average duration for novices than experts. These questions concerned post-processing, initialization, batching, end conditions for training termination and plotting, and required detailed code inspection.

\paragraph{LME.}
In order to analyze the dependence of scores and duration on particular questions and individual users, we performed a linear mixed effects (LME) analysis using the R library \texttt{lme4} \cite{LME4}.
Participants and questions are treated as random effects (categorical), the level of expertise as fixed effect (binary). Duration and score per question are response variables.\footnote{Modeling expertise with higher granularity instead of the binary classification into expertise groups (individual variables for experience with PyTorch, NMT and background in deep learning) did not have a significant effect on the model, since the number of participants is relatively low.}
For both response variables the variability is higher depending on the question than on the user (6x higher for score, 2x higher for time). The intercepts of the fixed effects show that novices score on average 0.14 points less while taking 2.47 min longer on each question than experts. 
The impact of the fixed effect is significant at $p<0.05$. 

\subsection{Findings}
First of all, we observe that the design of the questions was engaging enough for the students because all participants invested at least 1h to complete the quiz voluntarily. The experts also reported having gained new insights into the code through the quiz. We found that there are significant differences between both groups: Most prominently, the novices needed more time to answer each question, but still succeeded in answering the majority of questions correctly. There are larger variances within the group of novices, because they had to develop individual strategies to explore the code and use the available resources (documentation, code search, IDE), while experts could in many cases rely on prior knowledge.

\section{Conclusion}
We presented Joey NMT, a toolkit for sequence-to-sequence learning designed for NMT novices. It implements the most common NMT features and achieves performance comparable to more complex toolkits, while being minimalist in its design and code structure. In comparison to other toolkits, it is smaller in size and but more extensively documented. A user study on code accessibility confirmed that the code is comprehensibly written and structured. We hope that \joeynmt 
will ease the burden for novices to get started with NMT, 
and can serve as a basis for teaching.

\section*{Acknowledgments}
We would like to thank Sariya Karimova, Philipp Wiesenbach, Michael Staniek and Tsz Kin Lam for their feedback on the early stages of the code and for their bug fixes. We also thank the student and expert participants of the user study.

\newpage
\bibliography{references}
\bibliographystyle{acl_natbib_nourl}

\clearpage
\appendix
\onecolumn

\section{Supplemental Material}

\subsection{NMT Features}\label{app:features}
 Table~\ref{tab:feature-table} gives an overview over Joey NMT's features compared with several popular NMT toolkits implemented in Python, such as Sockeye, Neural Monkey, fair-seq, Tensor2Tensor (T2T), XNMT and OpenNMT-py. Sockeye is based on MXNet, Neural Monkey and Tensor2Tensor on TensorFlow, XNMT on Dynet and fair-seq, OpenNMT-py and JoeyNMT on PyTorch. We filled the table to our best knowledge with information obtained from GitHub repositories, published papers and provided documentation.

\subsection{Extra Results}

\paragraph{WMT14.}
WMT14 has been a popular benchmark to compare MT systems, even though different pre-/post-processing methods make comparisons noisy.\footnote{
See \url{https://github.com/tensorflow/tensor2tensor/issues/317} for a discussion on post-processing for en-de.}
We train a recurrent 1-layer (``shallow'') and  4-layer (``deep'') and a Transformer model on the same data as \citet{LuongETAL:15}.
Training the shallow RNN model took about 5 days on one P40 GPU; the deep model took around 9 days, the Transformer 10 days for en-de and 12 days for en-fr.
For comparative purposes we report (Moses-)tokenized and compound-splitted (only en-de) \texttt{multibleu} scores.
Table~\ref{tab:wmt14} compares the \joeynmt models against GNMT,  \citet{LuongETAL:15}, OpenNMT-py, and the original Tensor2Tensor Transformer. 
Without checkpoint averaging and extensive hyperparameter tuning, \joeynmt achieves results that come close to these systems.

\begin{table}[ht]
    \centering
    \begin{tabular}{lll}
         \toprule
          \textbf{System} & \textbf{en-de} & \textbf{en-fr}\\
         \midrule
         \citet{LuongETAL:15}  & 18.1 & 31.5\\
         GNMT & 24.6 & 39.0 \\ 
         \textbf{\joeynmt RNN} & 22.5 & 35.7 \\ 
        \textbf{\joeynmt RNN} (deep) & 24.0 & 37.4 \\ 
         \bottomrule
    \end{tabular}
    \caption{\texttt{newstest2014} results.}
    \label{tab:wmt14}
\end{table}
\paragraph{IWSLT En-Vi.}
We also compared our RNNs against Tensorflow NMT and XNMT on the IWSLT15 en-vi data set as pre-processed by Stanford. Table~\ref{tab:iwslt-vi} shows the results. The first three systems were trained on sentences of up to 50 tokens, while last two systems were trained on sentences of up to 110 tokens. Our BLEU scores were computed with SacreBLEU with version string BLEU+case.mixed+numrefs.1+smooth.exp+
tok.none+version.1.3.6. We use the original tokenization and data from \url{https://nlp.stanford.edu/projects/nmt}.

\begin{table}[h!]
    \centering
    \begin{tabular}{lc}
    \toprule
        \textbf{System} & \textbf{en-vi} \\
    \midrule
       \citet{LuongETAL:15} & 23.3 \\    
       TensorFlow NMT & 26.1\\
        \textbf{\joeynmt RNN} & 26.5	\\
        \midrule
        XNMT & 27.3 \\
        \textbf{\joeynmt RNN} & 27.7 \\
    \bottomrule
    \end{tabular}
    \caption{IWSLT en-vi}
    \label{tab:iwslt-vi}
\end{table}

\subsection{Crash Course on NMT and Pytorch Basics}\label{sec:course}
Prior to the study, the novices attended a three-day crash course (ca.~20h in total) where they were introduced to the concepts of feed-forward, recurrent and attentional neural networks, to PyTorch and the encoder-decoder model for sequence-to-sequence learning. In addition to lectures on the theory and background, they completed a subset of the PyTorch and RNN exercises of the Udacity course on Deep Learning\footnote{Parts 1-6 of the publicly available notebooks on \url{https://github.com/udacity/deep-learning-v2-pytorch/tree/master/intro-to-pytorch} and \url{https://github.com/udacity/deep-learning-v2-pytorch/tree/master/recurrent-neural-networks/char-rnn}, commit hash 9b6001a.}, so that they had all implemented and trained a feed-forward neural network for image classification and an LSTM for character-level language modeling. Solutions were discussed in class. In addition, they worked through the ``The Annotated Encoder Decoder''\footnote{ \url{https://github.com/bastings/annotated_encoder_decoder}} \cite{bastings2018annotated} to get a grasp of the building blocks of a NMT implementation in PyTorch. 
Note that we did not teach \joeynmt explicitly in class, but the students had to self-sufficiently work through a \joeynmt tutorial\footnote{\url{https://\joeynmt.readthedocs.io}}.

\subsection{Quiz Interface}
\begin{figure*}[h!]
    \centering
    \includegraphics[width=\textwidth]{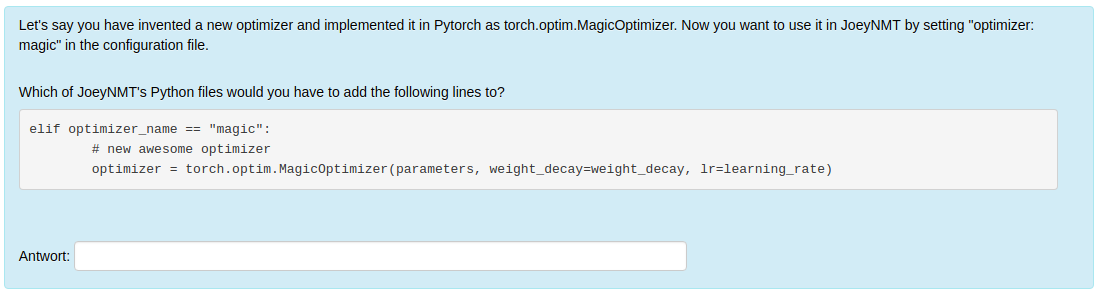}
    \includegraphics[width=\textwidth]{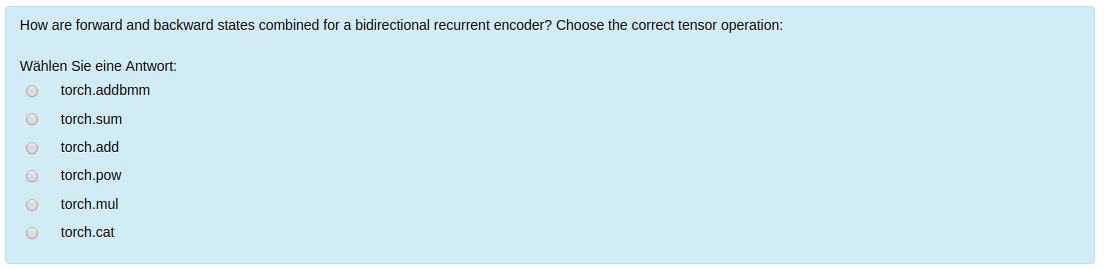}
    \caption{A free-text and a multiple-choice question. Instructions ''Antwort'' (response) are in German since the interface of the e-learning platform is.}
    \label{fig:questions}
\end{figure*}

Figure~\ref{fig:questions} shows the interface for two example questions, one as a free-text question, and one as a multiple-choice task.

\subsection{Quiz Statistics}\label{sec:stats}
\begin{figure}[h!]
    \centering
    \includegraphics[width=0.3\columnwidth]{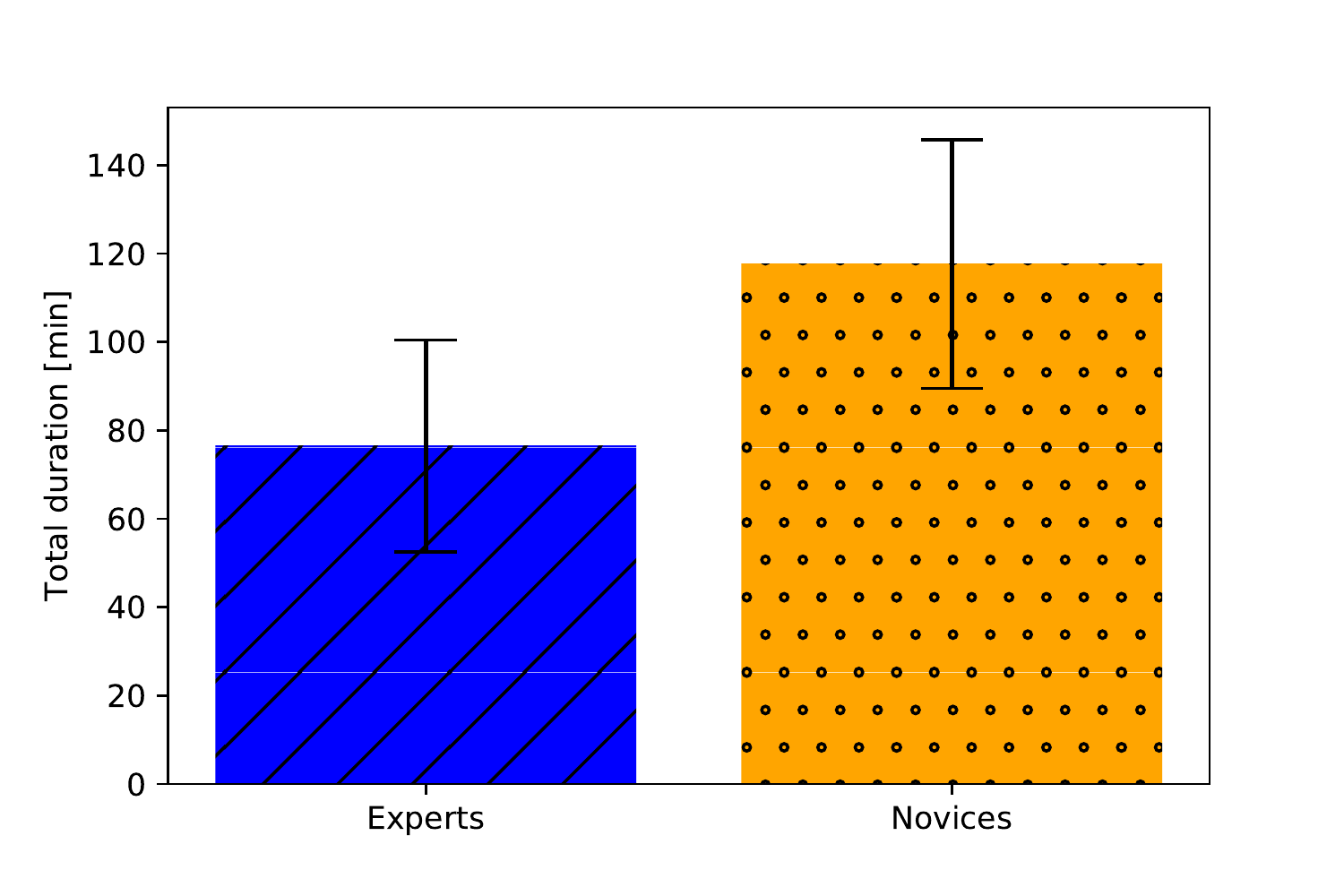}
    \caption{Total duration of quiz taken by experts and novices.}
    \label{fig:duration}
\end{figure}

\begin{figure}[h!]
    \centering
    \includegraphics[width=0.3\columnwidth]{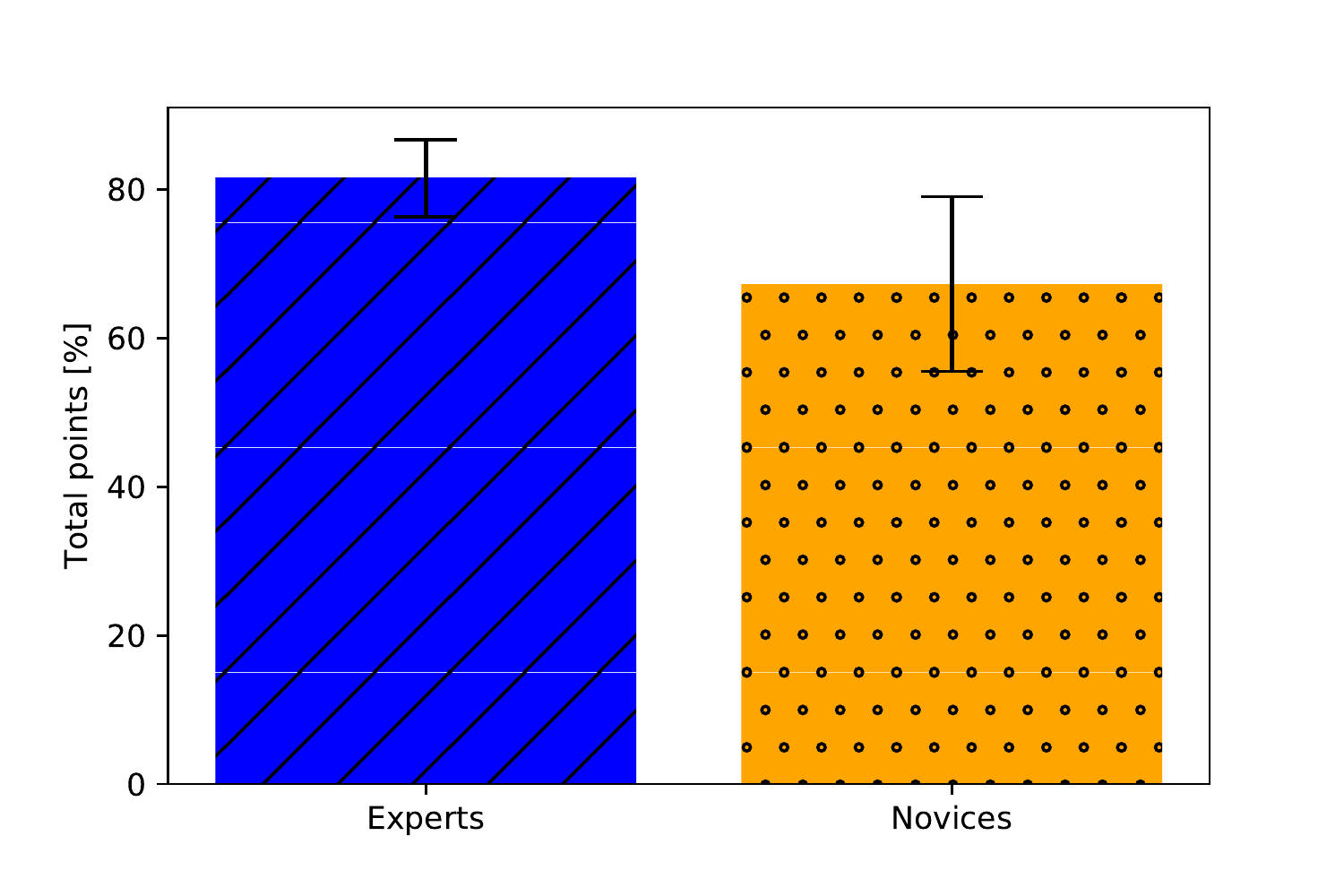}
    \caption{Percentage of points scored by experts and novices.}
    \label{fig:points}
\end{figure}

Figure~\ref{fig:duration} compares the total completion time for the quiz, Figure~\ref{fig:points} the total points between experts and novices.

\subsection{Quiz Questions}\label{sec:questions}
\begin{enumerate} 
\item \textbf{Training.} You have successfully installed \joeynmt and written a configuration file \texttt{config.yaml}. 
Which command would you use to start training a model with this configuration?
    \begin{itemize}
    \item \texttt{python3 -m joeynmt train config.yaml}
    \end{itemize}
\item \textbf{Translating.} Model training with \texttt{config.yaml} has finished and now you want to translate the pre-processed file \texttt{translate-me.txt} and save the translations in file \texttt{translated.txt} without specifying the file's path in the configuration file. Which command would you use?
    \begin{itemize}
    \item \texttt{python3   -m joeynmt translate config.yaml < translate-me.txt > translated.txt}
    \end{itemize}
\item \textbf{Saving.} How do you know your model was saved during training?
    \begin{itemize}
        \item[\cmark] Check in the validation report whether there's any line ending with "*".
        \item[\cmark] Check the training log if it says it saved checkpoints.
        \item[\cmark] Check if there are any \texttt{*.ckpt} files in the model directory.
        \item[\xmark] The model always gets saved during training.
    \end{itemize}
\item \textbf{Testing.} When using \joeynmt in test mode, can you specify the checkpoint for testing anywhere outside the configuration file?
    \begin{itemize}
        \item[\cmark] True
        \item[\xmark] False
    \end{itemize}
\item \textbf{Parameters.} How many parameters does the model specified in \texttt{configs/default.yaml} have in total? This includes all parameter weights and biases of the model, including e.g. the embeddings.
Hint: \joeynmt computes it for you.
    \begin{itemize}
        \item 66,376
    \end{itemize}
\item \textbf{Attention.} Which source token receives most attention when generating the target word ``if''?
\begin{figure}[h!]
    \centering
    \includegraphics[width=0.5\columnwidth]{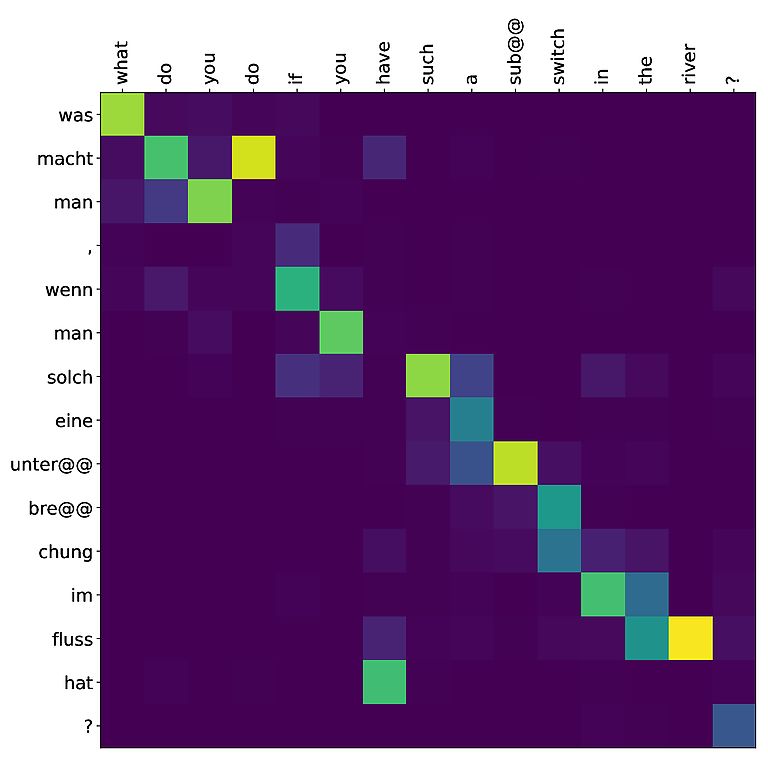}
\end{figure}
    \begin{itemize}
        \item ``wenn''
    \end{itemize}
\item \textbf{Speed.} How do you find out how fast your model trains (including validations)?
    \begin{itemize}
        \item The number of tokens per second is logged and reported in the log file.
    \end{itemize}
\item \textbf{Pre-processing.} Which pre-processing does \joeynmt do for you? (if specified)
    \begin{itemize}
        \item[\xmark] splitting into sub-word units (BPEs)
        \item[\xmark] data filtering by source/target length \mbox{ratio}
        \item[\cmark] data filtering by source and target length
        \item[\xmark] tokenization
        \item[\cmark] lowercasing
    \end{itemize}
\item \textbf{Post-processing.} Which post-processing does \joeynmt for you? (if specified)
    \begin{itemize}
        \item[\xmark] recasing
        \item[\xmark] detokenization
        \item[\cmark] subword merging (``un-BPE-ing'')
        \item[\xmark] delemmatization
    \end{itemize}
\item \textbf{Checkpoints.} In a debugging scenario, you don't want to store checkpoints for your current model.
There's a line that you can add to your configuration file to make the model not save any checkpoints during training. What is this line?
    \begin{itemize}
        \item \begin{verbatim} keep_last_ckpts: 0 \end{verbatim}
    \end{itemize}
\item \textbf{Model size.} Change the following model configuration to use three encoder layers. 
\begin{verbatim}
encoder:
  rnn_type: "lstm"
  embeddings:
    embedding_dim: 16
  hidden_size: 64
  bidirectional: True
\end{verbatim}
Which line would you have to add?
\begin{itemize}
    \item \begin{verbatim}num_layers: 3\end{verbatim}
\end{itemize}
\item \textbf{Data Path.} Which line would you have to add to the data configuration below to use \texttt{my\_home/my\_dir/my\_data.en} as test input file?
\begin{verbatim}
data:
  src: "en"
  trg: "fr"
  train: "test/data/reverse/train"
  dev: "test/data/reverse/dev"
  level: "word"
  lowercase: False
  max_sent_length: 25 
\end{verbatim}
Hint: mind the file ending.
\begin{itemize}
    \item \begin{verbatim}test: "my_home/my_dir/my_data"\end{verbatim}
\end{itemize}
\item \textbf{Training hyperparameters.} Modify the following training configuration such that it uses a constant learning rate of 0.02. 
\begin{verbatim}
training:
  optimizer: "adam"
  learning_rate: 0.001
  clip_grad_norm: 1.0
  batch_size: 10
  scheduling: "plateau"
  patience: 5
  decrease_factor: 0.5
  early_stopping_metric: "eval_metric"
  epochs: 6
  validation_freq: 1000
  logging_freq: 100
  model_dir: "reverse_model"
  max_output_length: 30  
\end{verbatim}
Paste the modified configuration below.
\begin{itemize}
    \item 
    \begin{verbatim}
training:
  optimizer: "adam"
  learning_rate: 0.02
  clip_grad_norm: 1.0
  batch_size: 10
  patience: 5
  decrease_factor: 0.5
  early_stopping_metric: "eval_metric"
  epochs: 6
  validation_freq: 1000
  logging_freq: 100
  model_dir: "reverse_model"
  max_output_length: 30
    \end{verbatim}
\end{itemize}
\item \textbf{Vocabulary generation.} When the vocabulary is extracted from training data, we keep only the \texttt{src\_voc\_limit / trg\_voc\_limit} most frequent tokens that occur at least \texttt{src\_min\_freq / trg\_min\_freq} times in the training data.

For the example, the vocabulary limit is 15, while the minimum frequency is 3.

After counting the tokens in the training data and filtering by minimum frequency, we have the following counts:
\begin{verbatim}
i: 22
you: 14
and: 9
,: 9
to: 7
if: 6
joey: 5
't: 5
anymore: 5
're: 4
scared: 4
be: 4
angry: 4
but: 3
it: 3
out: 3
don: 3
get: 3
oh: 3
the: 3
\end{verbatim}
Which of those tokens would \emph{not} end up in the vocabulary, according to \joeynmt's vocabulary building? 
\begin{itemize}
    \item[\cmark] it
    \item[\cmark] the
    \item[\cmark] oh
    \item[\cmark] get
    \item[\xmark] don
    \item[\xmark] but
    \item[\cmark] out
\end{itemize}
\item \textbf{Special tokens.} Which is the default token used for marking the end-of-sequence position in \joeynmt? e.g. \texttt{$<$end$>$} or \texttt{$[$EOS$]$}?
\begin{itemize}
    \item \begin{verbatim}</s>\end{verbatim}
\end{itemize}
\item \textbf{Data iterators.} Training and validation data are treated differently in \joeynmt - but in which ways?
For example, if you choose ``sorting'', it means that validation and training data are handled differently with respect to sorting - one gets sorted and the other doesn't.
\begin{itemize}
    \item[\cmark] Shuffling
    \item[\cmark] Filtering
    \item[\xmark] Tokenization
    \item[\xmark] Embedding
    \item[\cmark] Sorting
\end{itemize}
\item \textbf{Training loop.} Where is the training for-loop over epochs defined?
Paste the line in the textbox below. (Not the line number)
\begin{itemize}
    \item \begin{verbatim} for epoch_no in range(self.epochs):\end{verbatim} (in \texttt{training.py})
\end{itemize}
\item \textbf{End of training.} When does training end? (Assuming there are no technical problems like memory errors etc.)
We refer to settings in the configuration file, e.g. \texttt{learning\_rate}.
\begin{itemize}
    \item[\cmark] When the minimum learning rate (\texttt{learning\_rate\_min}) has been reached.
    \item[\xmark] When the maximum validation scores has been reached.
    \item[\xmark] Just after \texttt{keep\_last\_ckpts} checkpoints have been saved.
    \item[\xmark] When \joeynmt gets tired.
    \item[\cmark] When all epochs (\texttt{epochs}) have been completed.
    \item[\cmark] When you interrupt the training process with Ctrl+C.
\end{itemize}

\item \textbf{Model.} What does \texttt{model.forward()} return?
\begin{itemize}
 \item decoder outputs, decoder last hidden state, attention probabilities, attentional vectors
\end{itemize}
\item \textbf{Initialization.} How are forget gates of LSTMs initialized by default?
\begin{itemize}
 \item[\cmark] All ones
 \item[\xmark] Random normal initialization
 \item[\xmark] Random uniform initialization
\item[\xmark] All zeros
\item[\xmark] Xavier initialization
\end{itemize}
\item \textbf{Embeddings.} In the configuration we can "freeze" the embeddings, so that they are not (further) trained:
\begin{verbatim}
embeddings:
  embedding_dim: 16
  freeze: True
\end{verbatim}
Where does the freezing happen in JoeyNMT's code? Please give the freezing function's name.
\begin{itemize}
 \item \texttt{freeze\_params}
\end{itemize}
\item \textbf{Bidirectional.} How are forward and backward states combined for a bidirectional recurrent encoder? Choose the correct tensor operation:
\begin{itemize}
 \item[\xmark] \texttt{torch.add}
 \item[\cmark] \texttt{torch.cat}
 \item[\xmark] \texttt{torch.addbmm}
 \item[\xmark] \texttt{torch.sum}
 \item[\xmark] \texttt{torch.mul}
 \item[\xmark] \texttt{torch.pow}
\end{itemize}
\item \textbf{Bridge.} What's the name of the function that connects encoder and decoder by computing the initial decoder state given the last encoder state?
\begin{itemize} 
\item[\xmark] \texttt{bridge\_layer}
\item[\xmark] \texttt{BahdanauAttention}
\item[\xmark] \texttt{init\_decoder\_hidden}
\item[\xmark] \texttt{bridge\_layer}
\item[\xmark] \texttt{\_bridge}
\item[\xmark] \texttt{\_init\_decoder\_hidden}
\item[\xmark] \texttt{init\_hidden}
\item[\cmark] \texttt{\_init\_hidden}
\item[\xmark] \texttt{bridge}
\item[\xmark] \texttt{LuongAttention}
\item[\xmark] \texttt{\_attend}
\item[\xmark] \texttt{\_forward\_step}
\end{itemize}
\item \textbf{Loss computation.} Find the place where the batch loss is computed (comparing model outputs with targets), and paste the statement below.
e.g. \texttt{train\_batch\_loss = my\_loss\_function(outputs, targets)}
\begin{itemize}
 \item \begin{verbatim} batch_loss = loss_function(
    input=log_probs.contiguous().view(-1, log_probs.size(-1)), 
    target=batch.trg.contiguous().view(-1)) \end{verbatim}
\end{itemize}
\item \textbf{Batch.} During training, the Batch object in JoeyNMT holds the reference sequence in \texttt{trg} for computing the loss and in \texttt{trg\_input} for feeding it into the decoder.

What's the difference between those two tensors? (\texttt{batch.trg} vs. \texttt{batch.trg\_input})
\begin{itemize}
 \item[\xmark] $<$s$>$ is prepended to the first, otherwise no difference
 \item[\cmark] $<$/s$>$ is appended to the first and <s> is prepended to the second
 \item[\xmark] $<$s$>$ is prepended to the second, otherwise no difference
 \item[\xmark] $<$s$>$ is appended to the first and </s> is prepended to the second
 \item[\xmark] $<$/s$>$ is appended to the first, otherwise no difference
\end{itemize}
\item \textbf{Inference algorithm.} Where in the code is the decision made whether to decode greedily or with beam search? Paste the line below.

Hint: it's an if-statement.
\begin{itemize} 
 \item \texttt{if beam\_size == 0:}
\end{itemize} 
\item \textbf{Validation score computation.} Find the place where the validation score (here BLEU, \texttt{eval\_metric: bleu} is computed and paste the statement below.
\begin{itemize}
 \item \begin{verbatim}current_valid_score = bleu(valid_hypotheses, valid_references)\end{verbatim}
\end{itemize}
\item \textbf{BLEU computation.} Which library is used for BLEU score computation?
\begin{itemize}
 \item \texttt{sacrebleu}
\end{itemize}
\item \textbf{Optimizers.} Let's say you have invented a new optimizer and implemented it in Pytorch as \texttt{torch.optim.MagicOptimizer}. Now you want to use it in JoeyNMT by setting \texttt{optimizer: magic} in the configuration file.

Which of JoeyNMT's Python files would you have to add the following lines to?
\begin{verbatim}
elif optimizer_name == "magic":
    # new awesome optimizer
    optimizer=torch.optim.MagicOptimizer(
        parameters, weight_decay=weight_decay, lr=learning_rate)
\end{verbatim}
\begin{itemize}
 \item \texttt{builders.py}
\end{itemize}
\item \textbf{Attention.} For Bahdanau attention, find the line where the attention scores for a decoder hidden state are computed (before masking).
\begin{itemize}
 \item \begin{verbatim} scores = self.energy_layer(
    torch.tanh(self.proj_query + self.proj_keys))\end{verbatim}
\end{itemize}
\item \textbf{Plotting.} You want to use a different colormap for attention visualization, namely the one called \texttt{``binary''}. Give the line of JoeyNMT's code that is responsible for plotting the attention, modified to use the new colormap.
\begin{itemize}
 \item \begin{verbatim}plt.imshow(scores, cmap='binary', aspect='equal', 
    origin='upper', vmin=0., vmax=1.)\end{verbatim}
\end{itemize}
\end{enumerate}

\subsection{LMEM Details}\label{sec:lmem}
\subsection{Score per Question}
\begin{verbatim}
Linear mixed model fit by REML ['lmerMod']
Formula: score ~ group + (1 | item) + (1 | user)

Random effects:
 Groups   Name        Variance Std.Dev.
 item     (Intercept) 0.03668  0.1915  
 user     (Intercept) 0.00661  0.0813  
 Residual             0.12191  0.3492  
Number of obs: 434, groups:  item, 31; user, 14

Fixed effects:
            Estimate Std. Error t value
(Intercept)  0.81532    0.05422  15.036
groupnovice -0.14258    0.05545  -2.571

Correlation of Fixed Effects:
            (Intr)
groupnovice -0.584

all.model_score0: score ~ (1 | item) + (1 | user)
all.model_score: score ~ group + (1 | item) + (1 | user)
                 Df    AIC    BIC  logLik deviance  Chisq Chi Df Pr(>Chisq)  
all.model_score0  4 394.39 410.68 -193.20   386.39                           
all.model_score   5 390.50 410.87 -190.25   380.50 5.8904      1    0.01522 *
\end{verbatim}

\subsection{Time per Question}
\begin{verbatim}
Linear mixed model fit by REML ['lmerMod']
Formula: time ~ group + (1 | item) + (1 | user)

Random effects:
 Groups   Name        Variance Std.Dev.
 item     (Intercept)  0.8800  0.9381  
 user     (Intercept)  0.4834  0.6953  
 Residual             11.2855  3.3594  
Number of obs: 434, groups:  item, 31; user, 14

Fixed effects:
            Estimate Std. Error t value
(Intercept)   2.4677     0.4119   5.992
groupnovice   1.3266     0.4972   2.668

Correlation of Fixed Effects:
            (Intr)
groupnovice -0.690

all.model_time0: time ~ (1 | item) + (1 | user)
all.model_time: time ~ group + (1 | item) + (1 | user)
                Df    AIC    BIC  logLik deviance  Chisq Chi Df Pr(>Chisq)  
all.model_time0  4 2330.1 2346.4 -1161.0   2322.1                           
all.model_time   5 2325.7 2346.1 -1157.8   2315.7 6.3868      1     0.0115 *
\end{verbatim}

\begin{table*}[p]
    \centering
    \resizebox{\textwidth}{!}{
    \begin{tabular}{lccccccc}
         \toprule
         \textbf{Feature }& \textbf{Sockeye} & \textbf{Neural Monkey }& \textbf{fair-seq} &  \textbf{T2T }& \textbf{XNMT} & \textbf{OpenNMT-py} & \textbf{Joey NMT} \\
         \midrule
         \textit{Architecture} & & & & & & & \\
         RNN encoder            & \cmark & \cmark & \cmark & \cmark & \cmark & \cmark & \cmark \\
         RNN decoder            & \cmark & \cmark & \cmark & \cmark & \cmark & \cmark & \cmark \\
         Transformer encoder    & \cmark & \cmark & \cmark & \cmark & \cmark & \cmark & \cmark \\
         Transformer decoder    & \cmark & \cmark & \cmark & \cmark & \cmark & \cmark & \cmark \\
         ConvS2S encoder        & \cmark & \cmark & \cmark &        &        & \cmark & \\
         ConvS2S decoder        & \cmark &        & \cmark &        &        & \cmark & \\
         Image Encoder          & \cmark & \cmark &        & \cmark &        & \cmark & \\
         Audio Encoder          &        & \cmark & \cmark & \cmark & \cmark & \cmark & \\
         CTC                    &        & \cmark &        &        &        &        & \\
         Attention Mechanisms   & \cmark & \cmark & \cmark & \cmark & \cmark & \cmark & \cmark \\
         \midrule
         \textit{Tasks} & & & & & & & \\
         Embedding Tying        & \cmark &        & \cmark & \cmark & \cmark & \cmark & \cmark \\
         Softmax Tying          & \cmark & \cmark & \cmark & \cmark & \cmark & \cmark & \cmark \\ 
         Parameter Freezing     & \cmark &        & \cmark &        &        &        & \cmark \\
         Multi-Source           &        & \cmark & \cmark &        & \cmark &        & \\
         Factored Input         & \cmark & \cmark &        &        &        & \cmark & \\
         Multi-Task             &        & \cmark & \cmark &        & \cmark  &        & \\
         Sequence Labeling      &        & \cmark &        &        &        &        & \\
         Sequence Classification&        & \cmark &        & \cmark &        &        & \\
         Language Modeling      &        & \cmark  & \cmark & \cmark &  \cmark & \cmark  & \\
         \midrule
         \textit{Inference} & & & & & & & \\
         Segmentation Levels (word/char/bpe) &\cmark & \cmark & \cmark & \cmark & \cmark & \cmark & \cmark \\ 
         Beam Search            & \cmark & \cmark & \cmark & \cmark & \cmark & \cmark & \cmark \\ 
         n-best outputs         & \cmark & \cmark & \cmark &        &      &     & \\
         Sampling               & \cmark & \cmark     & \cmark &        & \cmark &\cmark &   \\
         Rescoring              & \cmark & \cmark &        &        & \cmark &  \\
         Checkpoint averaging   & \cmark & \cmark & \cmark &  \cmark  & \cmark & \cmark & \cmark\\
         \midrule
         \textit{Training} & & & & & & & \\
         MLE                    & \cmark & \cmark & \cmark & \cmark & \cmark & \cmark & \cmark \\
         MRT                    &        & \cmark &        &        & \cmark && \\
         Gradient Clipping      & \cmark & \cmark & \cmark & \cmark & \cmark & \cmark & \cmark \\
         Dropout                & \cmark & \cmark & \cmark & \cmark & \cmark & \cmark & \cmark \\
         Weight Decay           & \cmark & \cmark & \cmark & \cmark & \cmark & \cmark & \cmark \\
         Label Smoothing        & \cmark & \cmark & \cmark & \cmark & \cmark & \cmark & \cmark \\
         Optimizer              & \cmark & \cmark & \cmark & \cmark & \cmark & \cmark & \cmark \\
         Scheduler              & \cmark & \cmark & \cmark & \cmark & \cmark & \cmark & \cmark \\ 
         Early Stopping         & \cmark & \cmark & \cmark & \cmark & \cmark & \cmark & \cmark \\ 
         \midrule
         \textit{Usage} & & & & & & & \\
         CPU/GPU                & \cmark & \cmark & \cmark & \cmark & \cmark & \cmark & \cmark \\
         Monitoring             & \cmark & \cmark & \cmark & \cmark & \cmark & \cmark & \cmark \\ 
         Attention Visualization& \cmark & \cmark && \cmark &\cmark&& \cmark \\
         \bottomrule
    \end{tabular}
    }
    \caption{Features implemented by popular NMT toolkits in Python as of July 1, 2019.}
    \label{tab:feature-table}
\end{table*}


\end{document}